\documentclass{article}

\PassOptionsToPackage{numbers, compress}{natbib} 
\usepackage[final]{src/neurips_data_2024}

\usepackage[utf8]{inputenc} %
\usepackage[T1]{fontenc}    %
\usepackage{hyperref}       %
\usepackage{url}            %
\usepackage{booktabs}       %
\usepackage{amsfonts}       %
\usepackage{nicefrac}       %
\usepackage{microtype}      %
\usepackage{xcolor}         %
\usepackage{colortbl}
\usepackage{graphicx}
\usepackage{amsmath,amsfonts,amssymb,amsthm}
\usepackage{color}
\usepackage{caption}
\usepackage{subcaption}
\usepackage{xspace}
\usepackage{array}
\usepackage{cite}
\usepackage{multirow}
\usepackage{pifont}
\usepackage{wrapfig}
\usepackage{adjustbox}

\DeclareUrlCommand\url{\color{ellisred}}
\hypersetup{
  colorlinks=true,
  urlcolor=ellisred, 
  linkcolor=ellisred,
  citecolor=ellisblue,
  pdfborder={0 0 0},
}

\title{NAVSIM: Data-Driven Non-Reactive \\ Autonomous Vehicle Simulation and Benchmarking}

\vspace{-0.0cm}
\author{
Daniel Dauner$^{1,2}$ \quad
Marcel Hallgarten$^{1,5}$ \quad
Tianyu Li$^{3}$ \quad
Xinshuo Weng$^{4}$ \quad
\textbf{Zhiyu Huang}$^{4,6}$ \\
\textbf{Zetong Yang}$^{3}$ \quad
\textbf{Hongyang Li}$^{3}$ \quad
\textbf{Igor Gilitschenski}$^{7,8}$ \quad
\textbf{Boris Ivanovic}$^{4}$ \quad
\textbf{Marco Pavone}$^{4,9}$ \\
\textbf{Andreas Geiger}$^{1,2}$ \quad
\textbf{Kashyap Chitta}$^{1,2}$
\vspace{0.0cm}
\\
$^{1}$University of T{\"u}bingen \quad
$^{2}$T{\"u}bingen AI Center \quad
$^{3}$OpenDriveLab at Shanghai AI Lab \\
$^{4}$NVIDIA Research \quad
$^{5}$Robert Bosch GmbH \quad
$^{6}$Nanyang Technological University \\
$^{7}$University of Toronto \quad
$^{8}$Vector Institute \quad
$^{9}$Stanford University
\vspace{-0.1cm}
}

\newcommand{\figref}[1]{Fig.~\ref{#1}}
\newcommand{\secref}[1]{Section~\ref{#1}}

\newcommand{\tabref}[1]{Table~\ref{#1}}

\makeatletter
\DeclareRobustCommand\onedot{\futurelet\@let@token\@onedot}
\def\@onedot{\ifx\@let@token.\else.\null\fi\xspace}
\def\eg{e.g\onedot} 
\def\ie{i.e\onedot}

\makeatother

\newcommand{\boldparagraph}[1]{\vspace{0.0cm}\noindent{\bf #1.}}

\definecolor{darkgreen}{rgb}{0,0.7,0}
\definecolor{darkyellow}{rgb}{0.8,0.8,0}
\definecolor{bittersweet}{rgb}{1.0, 0.44, 0.37}
\definecolor{amber}{rgb}{1.0, 0.49, 0.0}
\definecolor{lgray}{rgb}{0.83,0.83,0.83}

\definecolor{color_unlabled}{rgb}{0.0,0.0,0.0}
\definecolor{color_vehicle}{rgb}{0.0,0.0,0.56}
\definecolor{color_road}{rgb}{0.5,0.25,0.5}
\definecolor{color_redlight}{rgb}{1.0,0.0,0.0}
\definecolor{color_person}{rgb}{0.859,0.078,0.234}
\definecolor{color_roadline}{rgb}{0.613,0.914,0.195}
\definecolor{color_sidewalk}{rgb}{0.953,0.137,0.906}

\definecolor{ellisred}{rgb}{0.87,0.44,0.38} %
\definecolor{ellisgreen}{rgb}{0.69,0.90,0.52} %
\definecolor{elliscyan}{rgb}{0.29,0.77,0.74} %
\definecolor{ellisorange}{rgb}{0.89,0.55,0.28} %
\definecolor{ellisblue}{rgb}{0.41,0.61,0.86} %

\definecolor{Tab0}{HTML}{1F77B4}
\definecolor{Tab1}{HTML}{ff7f0e}
\definecolor{Tab2}{HTML}{2ca02c}
\definecolor{Tab3}{HTML}{d62728}
\definecolor{Tab4}{HTML}{9467bd}
\definecolor{Tab5}{HTML}{8c564b}
\definecolor{Tab6}{HTML}{e377c2}
\definecolor{Tab7}{HTML}{7f7f7f}
\definecolor{Tab8}{HTML}{bcbd22}
\definecolor{Tab9}{HTML}{17becf}

\definecolor{Tabx0}{HTML}{4e79a7}
\definecolor{Tabx1}{HTML}{f28e2b}
\definecolor{Tabx2}{HTML}{e15759}
\definecolor{Tabx3}{HTML}{76b7b2}
\definecolor{Tabx4}{HTML}{59a14f}
\definecolor{Tabx5}{HTML}{edc948}
\definecolor{Tabx6}{HTML}{b07aa1}
\definecolor{Tabx7}{HTML}{ff9da7}
\definecolor{Tabx8}{HTML}{9c755f}
\definecolor{Tabx9}{HTML}{bab0ac}

\newcommand{\cmark}{\ding{51}}

\begin{document}

\maketitle

\begin{abstract}
Benchmarking vision-based driving policies is challenging. On one hand, open-loop evaluation with real data is easy, but these results do not reflect closed-loop performance. On the other, closed-loop evaluation is possible in simulation, but is hard to scale due to its significant computational demands. Further, the simulators available today exhibit a large domain gap to real data. This has resulted in an inability to draw clear conclusions from the rapidly growing body of research on end-to-end autonomous driving. In this paper, we present NAVSIM, a middle ground between these evaluation paradigms, where we use large datasets in combination with a non-reactive simulator to enable large-scale real-world benchmarking. Specifically, we gather simulation-based metrics, such as progress and time to collision, by unrolling bird's eye view abstractions of the test scenes for a short simulation horizon. Our simulation is non-reactive, \textit{i.e.}, the evaluated policy and environment do not influence each other. As we demonstrate empirically, this decoupling allows open-loop metric computation while being better aligned with closed-loop evaluations than traditional displacement errors. NAVSIM enabled a new competition held at CVPR 2024, where 143 teams submitted 463 entries, resulting in several new insights. On a large set of challenging scenarios, we observe that simple methods with moderate compute requirements such as TransFuser can match recent large-scale end-to-end driving architectures such as UniAD. Our modular framework can potentially be extended with new datasets, data curation strategies, and metrics, and will be continually maintained to host future challenges. Our code is available at  \href{https://github.com/autonomousvision/navsim}{{\color{ellisred}{\texttt{https://github.com/autonomousvision/navsim}}}}.
\end{abstract}

\begin{figure}[htbp]
\vspace{-0.0cm}
\centering
\includegraphics[width=\textwidth]{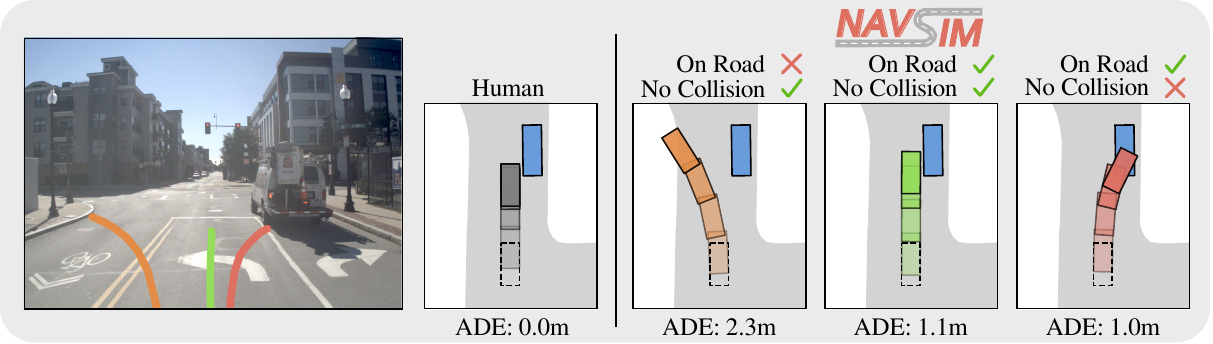} \\
\vspace{-0.0cm}
\caption{\textbf{NAVSIM.} Traditional metrics such as the average displacement error (ADE) overlook the multi-modality of driving. They penalize trajectories that deviate from a recorded human driving log, even if such a trajectory is safe. Our benchmark evaluates trajectory outputs of sensor-based driving policies with simulation-based metrics, considering collisions and map compliance.
}
\label{fig:teaser}
\vspace{0.1cm}
\end{figure}

\section{Introduction}

Autonomous vehicles (AVs) have gained immense research interest due to their potential to change transportation and improve traffic safety~\citep{Janai2020,Chen2023ARXIV}. This has created a large community working on the development of AV algorithms, which map high-dimensional sensor data to desired vehicle control outputs. Therefore, measuring and comparing the performance of AV algorithms is a crucial task.

Unfortunately, it is extremely challenging to evaluate driving performance, and the most widely-used benchmarks today fall short in several respects: (1) the datasets used, such as nuScenes~\citep{Caesar2021CVPR}, were created for perception tasks such as object detection. As such, they focus on visual diversity and label quality instead of the relevance of the data for research on planning. Often, most frames have a trivial solution of extrapolating the historical driving behavior, leading to ``blind'' driving policies that observe only the vehicle's past trajectory obtaining state-of-the-art performance
~\citep{Zhai2023ARXIV, Li2024CVPR, Dauner2023CORL}. (2) Due to the fact that driving is an inherently multifaceted task where the algorithm must coordinate several desired properties such as safety, comfort, and progress, the evaluation metric must also balance potentially conflicting objectives. However, as shown in \figref{fig:teaser}, existing metrics such as the average displacement error (ADE) between a predicted and recorded human trajectory often misrepresent the relative accuracy of trajectories. (3) Since driving involves interactions among multiple agents, evaluation must ideally be interactive, \eg, in simulation. Unfortunately, existing simulators with synthetic sensor data exhibit a significant domain gap to real-world driving. (4) Besides, the lack of a standardized evaluation setup has led to subtle inconsistencies between metrics in existing work, leading to unfair comparisons and inaccurate conclusions~\citep{Weng2024CVPR, Li2024CVPR}. Collectively, these problems hinder progress in the development of AVs, emphasizing the need for more principled benchmarks.

In this work, we take steps towards alleviating these issues. First, we propose a strategy for sampling interesting driving scenarios and apply it to the largest publicly-available driving dataset~\citep{Karnchanachari2024ICRA}. We obtain, for the first time, over 100k challenging real-world driving scenarios for training and evaluating sensor-based driving policies. We show that in these scenarios, ``blind'' driving policies fail to compete with more principled sensor-based policies. Second, we draw inspiration from the literature of rule-based planning for AVs~\citep{Sauer2018CORL,Fan2018ARXIV,Sadat2019IROS,Dauner2023CORL} to identify a set of diverse, efficient, and principled metrics that cover multiple facets of the autonomous driving task. Third, we circumvent the need for inaccurate sensor simulation with domain gaps by simplifying our simulation to a non-reactive one. Given an observed real-world sensor input, the agent under test commits to a set of actions for a specific time horizon. Further, these actions are assumed to not affect the future behavior of other agents in the scene. Under this setting, it is possible to simulate the expected motion of all agents over this time horizon in a simplified bird's-eye-view (BEV) abstraction of the scene, and incorporate metrics that involve interactions, as we observe in \figref{fig:teaser}. Empirically, we demonstrate that our selected metrics are well-correlated to the outcomes of closed-loop simulations. Finally, we establish an official evaluation server on the open-source HuggingFace platform, which is free, has a low maintenance overhead, and enables future scaling to more challenging datasets and metrics.

We combine these ideas to propose NAVSIM, a comprehensive tool for AV data curation, simulation, and benchmarking. We instantiate standardized training and evaluation splits for NAVSIM with the OpenScene dataset~\citep{OpenScene2023}, though our framework can be extended to other datasets. With these splits, we present a detailed analysis of popular end-to-end driving models previously benchmarked either exclusively on CARLA~\citep{Dosovitskiy2017CORL} or nuScenes~\citep{Caesar2021CVPR}, providing the first direct comparison between these families of approaches in an independent evaluation setting. Interestingly, we find that the performances of the best methods developed in both settings are similar, despite a vast difference in computational requirements for their training. Finally, we review the insights gained through the 2024 NAVSIM challenge\footnote{\href{https://opendrivelab.com/challenge2024/\#end\_to\_end\_driving\_at\_scale}{{\color{ellisred}{\texttt{https://opendrivelab.com/challenge2024/\#end\_to\_end\_driving\_at\_scale}}}}}, hosted in conjunction with the CVPR 2024 Workshop on Foundation Models for Autonomous Systems. For the challenge, 143 teams from 13 countries developed diverse methods that competed on the proposed benchmark. The top methods ranged from multi-billion parameter vision language models~\citep{Sima2023ARXIV,Ma2023ARXIV,Yang2023ARXIV,Zhou2024ARXIV} to more efficient and recently overlooked approaches based on trajectory sampling and scoring~\citep{Sadat2020ECCV,Hu2022ECCV,Chen2024ARXIV}, demonstrating the remarkable ability of the broader community to advance AV research when provided with the right tools.

\boldparagraph{Contributions} (1) We build NAVSIM, a framework for \underline{n}on-reactive \underline{AV} \underline{sim}ulation, with standardized protocols for training and testing, data curation tools ensuring broad accessibility, and an official public evaluation server used for the inaugural NAVSIM challenge. (2) We develop configurable simulation-based metrics that are well-suited for evaluating sensor-based motion planning. (3) We reimplement a collection of end-to-end approaches for NAVSIM including TransFuser, UniAD, and PARA-Drive, showcasing the surprising potential of simple models in our challenging scenarios.
\section{Related Work}

\boldparagraph{End-to-End Driving} End-to-end driving streamlines the entire stack from perception to planning into a single optimizable network. This eliminates the need for manually designing intermediate representations. Following pioneering work~\citep{Pomerleau1988NIPS, Bojarski2016ARXIV, Kendall2018ARXIV}, a diverse landscape of end-to-end models has emerged. For instance, an extensive body of end-to-end approaches focuses on closed-loop simulators, utilizing single-frame cameras, LiDAR point clouds, or a combination of both for expert imitation~\citep{Chen2019CORL, Chitta2021ICCV, Prakash2021CVPR, Chen2021ICCVa, Wu2022NeurIPS, Shao2022CORL, Shao2023CVPR, Chitta2023PAMI, Jia2023CVPR, Zhang2023CVPR, Jaeger2023ICCV}. More recently, developing end-to-end models on open-loop benchmarks has gained traction~\citep{Hu2022ECCV, Hu2023CVPR, Jiang2023ICCV, Ye2023ARXIV, Li2024CVPR, Weng2024CVPR}. Our work introduces a new evaluation scheme with which we compare end-to-end models from both communities. 

\boldparagraph{Closed-Loop Benchmarking with Simulation} Driving simulators allow us to evaluate autonomous systems in a closed-loop manner and collect downstream driving statistics, including collision rates, traffic-rule compliance, or comfort. A broad body of research conducts evaluations in simulators, such as CARLA~\citep{Dosovitskiy2017CORL} or Metadrive~\citep{Li2022PAMI} with sensor simulation, or nuPlan~\citep{Karnchanachari2024ICRA} and Waymax~\citep{Gulino2023NIPS} for data-driven simulation. Unfortunately, ensuring realism when simulating traffic behavior or sensor data remains a challenging task. To simulate camera or LiDAR sensors, most established simulators rely on graphics-based rendering methods, leading to an inherent domain gap in terms of visual fidelity and sensor characteristics. Data-driven simulators for motion planning incorporate traffic recordings but do not support image or LiDAR-based methods~\citep{Karnchanachari2024ICRA, Gulino2023NIPS, Chitta2024ARXIV}. Data-driven sensor simulation leverages and adapts real-world sensor data to create new simulations where the vehicle may move differently, but the rendering quality of existing tools is subpar~\citep{Amini2020RAL, Amini2022ICRA, Wang2022ICRA}. Further, while promising image~\citep{Tonderski2024CVPR} or LiDAR~\citep{Manivasagam2023ICCV} synthesis approaches exist, efficiently simulating sensors entirely from data remains an open problem. In this work, we provide an approach for the evaluation of real sensor data with simulation-based metrics by making a simplifying assumption that the agent and environment do not influence each other over a short simulation horizon. Despite this strong assumption, when benchmarking on real data, NAVSIM better reflects planning performance than established evaluation protocols, as demonstrated through our systematic experimental analysis.

\boldparagraph{Open-Loop Benchmarking with Displacement Errors} Open-loop evaluation protocols commonly measure displacement errors between trajectories of a recorded expert (i.e., of a human driver) and a motion planner. However, several issues concerning evaluation with displacement errors have surfaced recently, particularly on the nuScenes dataset~\citep{Caesar2021CVPR}. Given that nuScenes does not provide standardized planning metrics, prior work relied on independent implementations, which led to inconsistencies when reporting or comparing results~\citep{Weng2024CVPR, Li2024CVPR}. Next, most planning models in nuScenes receive the human trajectory endpoint as a discrete direction command~\citep{Hu2022ECCV, Hu2023CVPR, Jiang2023ICCV, Li2024CVPR, Weng2024CVPR}, thereby leaking ground-truth information into inputs. Moreover, about 75\% of the scenarios in nuScenes involve trivial straight driving~\citep{Li2024CVPR}, leading to simple solutions when extrapolating the ego-motion. For instance, AD-MLP demonstrates that an MLP on the kinematic ego status (ignoring perception completely) can achieve state-of-the-art displacement errors~\citep{Zhai2023ARXIV}. Such blind agents are undeniably dangerous, which highlights a broader concern: displacement metrics are not correlated to closed-loop driving~\citep{Codevilla2018ECCV, Dosovitskiy2017CORL, Bansal2019RSS, Dauner2023CORL}. In this work, we address prevalent issues of nuScenes and propose a standardized driving benchmark with challenging scenarios and an official evaluation server. We derive a navigation goal from the lane graph instead of the human trajectory to prevent label leakage, and propose principled simulation-based metrics as an alternative to displacement errors.
\section{NAVSIM: Non-Reactive Autonomous Vehicle Simulation}
\label{sec:3}

NAVSIM combines the ease of use of open-loop benchmarks such as nuScenes~\citep{Caesar2021CVPR} with metrics based on closed-loop simulators such as nuPlan~\citep{Karnchanachari2024ICRA}.  In the following, we give a detailed introduction to the task and metrics that driving agents are challenged with in NAVSIM. Subsequently, we propose a filtering method to obtain standardized train and test splits covering challenging scenes.

\boldparagraph{Task description} Driving agents in NAVSIM must plan a trajectory, defined as a sequence of {future} poses, over a horizon of $h$ seconds. Their input contains streams of \textit{past} frames from onboard sensors, such as cameras, LiDAR, as well as the vehicle's current speed, acceleration, and navigation goal, jointly termed the ego status. For compatibility with prior work~\citep{Hu2022ECCV, Hu2023CVPR, Jiang2023ICCV, Weng2024CVPR}, we provide the navigation goal as a one-hot vector with three categories: left, straight, or right. 

\boldparagraph{Non-Reactive Simulation} Traditional closed-loop benchmarks normally infer planners at high frequencies, \eg, 10Hz~\citep{Dosovitskiy2017CORL,Karnchanachari2024ICRA}. However, this requires efficient simulation of all input modalities for the driving agent, including high-dimensional sensor streams in the case of sensor-based approaches. To sidestep this, the core idea of NAVSIM is to evaluate driving agents using a non-reactive simulation. This means driving agents are only queried in the initial frame of each scene. Afterwards, the planned trajectory is kept fixed for the entire trajectory duration. Over this short horizon, no environmental feedback is provided to the driving agent, and the NAVSIM evaluation is purely based on the initial real-world sensor sample. This makes the agent's task more challenging, limiting simulations to short horizons. We select a horizon of $h=4$ seconds, which has been shown in prior work to be adequate for closed-loop planning~\citep{Dauner2023CORL}. Despite this limitation, non-reactive simulation offers a key advantage: unlike traditional open-loop benchmarks, which mainly compare the planned trajectory to the human driver's trajectory in a similar setting, it enables the use of simulation outcomes to compute {metrics reflecting safety, comfort, and progress}. An LQR controller~\citep{Lehtomaki1981TAC} is applied at each simulation iteration to calculate steering and acceleration values, and a kinematic bicycle model~\citep{Rajamani2011Springer} propagates the ego vehicle. We execute this pipeline at $10$Hz over the $4$s trajectory horizon. In Sec.~\ref{sec:4_alignment}, we show that despite our simplifying assumption, our evaluation results in a much better alignment with closed-loop metrics than traditional open-loop metrics achieve.

\boldparagraph{PDM Score} NAVSIM scores driving agents in two steps. First, subscores in range $[0,1]$ are computed after simulation. Second, these subscores are aggregated into the PDM Score (PDMS) $\in[0,1]$. It is named after the Predictive Driver Model (PDM)~\citep{Dauner2023CORL}, a state-of-the-art rule-based planner which uses this scoring function to evaluate trajectory proposals during closed-loop simulation in nuPlan. The metric is also an efficient reimplementation of the nuPlan closed-loop score metric~\citep{Karnchanachari2024ICRA}. In NAVSIM, the PDMS can be adapted by adding or removing subscores, changing aggregation parameters, or making subscores more challenging, e.g., by adapting their internal thresholds. It is calculated per frame and averaged across frames. In this work, we use the following aggregation of subscores:
\begin{equation}
        \textrm{PDMS} = \underbrace{\Bigg( { \prod_{m \in \{\texttt{NC}, \texttt{DAC}\}}} \texttt{score}_m \Bigg)}_{\text{penalties}} \times \underbrace{\Bigg( \frac{\sum_{w \in \{\texttt{EP}, \texttt{TTC}, \texttt{C}\}} \texttt{weight}_w \times  \texttt{score}_w}{\sum_{w \in \{\texttt{EP}, \texttt{TTC}, \texttt{C}\}} \texttt{weight}_w }  \Bigg)}_{\text{weighted average}}.
\end{equation}
Subscores are categorized by their importance as penalties or terms in a weighted average. A penalty punishes inadmissible behavior such as collisions with a factor $<1$. The weighted average aggregates subscores for other objectives such as progress and comfort. In the following, we briefly describe each subscore. More details can be found in the supplementary material.

\boldparagraph{Penalties} Avoiding collisions and staying on the road is imperative for motion planning as it ensures traffic rule compliance and the safety of pedestrians and road users.
Thus, failing to drive with no collisions (NC) with road users (vehicles, pedestrians, and bicycles) or infractions with regard to drivable area compliance (DAC) result in hard penalties of $\texttt{score}_\texttt{NC}=0$ or $\texttt{score}_\texttt{DAC}=0$ respectively. This results in a PDMS of 0 for the current scene. We ignore certain collisions that are not considered "at-fault" in the non-reactive environment, \eg when the ego vehicle is static. For collisions with static objects, we apply a softer penalty of $\texttt{score}_\texttt{NC}=0.5$.

\boldparagraph{Weighted Average} The weighted average accounts for ego progress (EP), time-to-collision (TTC), and comfort (C). The ego progress subscore $\texttt{score}_\texttt{EP}$ represents the agent progress along the route center as a ratio to an approximated safe upper bound from the PDM-Closed planner~\citep{Dauner2023CORL}. PDM-Closed obtains a possible progress value without collisions or off-road driving with a search-based strategy based on trajectory proposals. The final ratio is clipped to $[0,1]$ while discarding low or negative progress scores if the upper bound is below 5 meters. Next, the TTC subscore ensures that driving agents respect the safety margins to other vehicles. Defaulting to a value of $1$, this subscore is set to $0$ if for any simulation step within the $4\text{s}$ horizon, the ego-vehicle's time-to-collison, when projected forward with a constant velocity and heading, is less than a certain threshold. Finally, the comfort subscore is obtained by comparing the acceleration and jerk of the trajectory to predetermined thresholds. Following the cost weights used by the PDM-Closed planner and the 2023 nuPlan Challenge, we set the coefficients of the weighted average as $\texttt{weight}_\texttt{EP}=5$, $\texttt{weight}_\texttt{TTC}=5$, and $\texttt{weight}_\texttt{C}=2$. We find this selection reasonable and robust to changes. For example, the top 3 ranks of the NAVSIM challenge remain identical when assigning an equal weight to the subscores.

\subsection{Generating Standardized and Challenging Train and Test Splits}
\label{sec:3_splits}

\boldparagraph{Dataset} The NAVSIM framework is agnostic to the choice of driving dataset. We choose OpenScene~\citep{OpenScene2023}, a redistribution of nuPlan~\citep{Karnchanachari2024ICRA}, the largest annotated public driving dataset. OpenScene includes 120 hours of driving at a reduced frequency of $2$Hz typically considered by end-to-end planning algorithms, resulting in a $90\%$ reduction of data storage requirements compared to nuPlan from over 20 TB to 2 TB. Our agent input, based on OpenScene, comprises eight cameras, each with a resolution of $1920\times1080$ pixels, and a merged LiDAR point cloud from five sensors. The input includes the current time-step and optionally 3 past frames, totaling $1.5$s at $2$Hz. In principle, any driving dataset that provides annotated HD maps, object bounding boxes, and sensor data can be converted into this format and thus be used with NAVSIM. 

\begin{figure}[t!]
\centering
\includegraphics[width=1.0\textwidth]{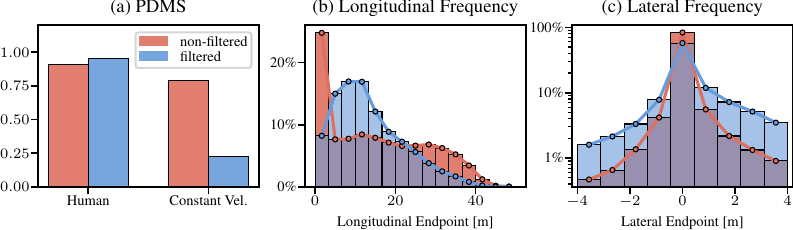} 
\caption{\textbf{Filtering.} (a) We consider challenging scenes where maintaining a constant velocity and heading fails compared to the human driver. (b) Our filtering primarily removes scenes with static or fast longitudinal movement and (c) leads to more diversity in lateral movement (log-scale).}
\label{fig:filtering}
\vspace{-0.0cm}
\end{figure}

\boldparagraph{Filtering for challenging scenes} A majority of human driving data involves trivial situations such as being stationary or straight driving at a near constant speed. These can be solved efficiently by simple heuristics, \eg, as depicted in Fig.~\ref{fig:filtering} (\hyperref[fig:filtering]{a}), the baseline of maintaining a constant velocity and heading achieves a PDMS of $79$\% on the OpenScene dataset, where human-level performance corresponds to $91$\%. In NAVSIM, we propose the use of a filtered dataset to remove frames with (1) near-trivial solutions and (2) significant annotation errors. We remove highly simplistic scenes by detecting if the previously mentioned constant velocity agent exceeds a PDMS of $0.8$. Similarly, we remove scenes in which the human trajectory results in a PDMS of less than $0.8$. This ensures that an acceptable solution exists to these difficult scenarios and filters out noisy annotations such as inaccurate bounding boxes. These thresholds can be adjusted based on the desired filtered dataset size. The resulting scenarios are challenging, which is underlined by the score of the constant velocity agent dropping to $22$\%, whereas the human expert achieves a score of $95$\%. The higher ratio of non-trivial scenarios, such as turning, also results in endpoints being less distant longitudinally when nonzero, and more evenly distributed laterally, as seen in Fig.~\ref{fig:filtering} (\hyperref[fig:filtering]{b-c}). We employ this filtering strategy to provide standardized splits for training and testing, called \texttt{navtrain} and \texttt{navtest}, with 103k and 12k samples respectively. This curated data serves as a benchmark accessible as a standalone download option with a moderate storage demand given its large scale and diversity (450 GB).
\section{Experiments}
\label{sec:4_experiments}

In this section, we present the results of our experiments aimed at answering the following questions: (1) Can non-reactive open-loop simulation provide sufficient correlation to closed-loop metrics? (2) What new conclusions do experiments on NAVSIM provide compared to prior benchmarks?

\begin{figure}[t!]
    \centering
    \includegraphics[width=1.0\textwidth]{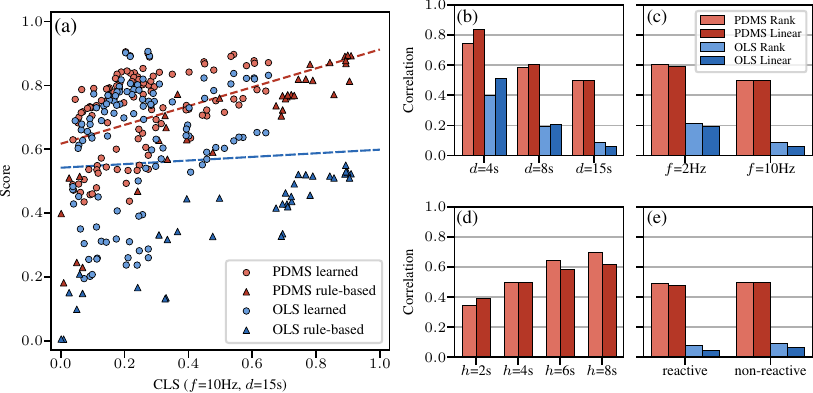}
    \caption{\textbf{Closed-Loop Alignment.} (a) For each planner, we show open-loop metrics (OLS, PDMS) together with the corresponding closed-loop score (CLS). The trendlines depicting correlations are fit linearly to all (learned and rule-based) planners. Moreover, we analyze different (b) CLS durations $d$, (c) planning frequencies $f$, (d) PDMS horizons $h$, and (e) closed-loop background agent behaviors. 
    }
    \label{fig:correlation}
    \vspace{-0.3cm}
\end{figure}

\subsection{Alignment Between Open-Loop and Closed-Loop Evaluation}
\label{sec:4_alignment}

Open-loop metrics should ideally be aligned with closed-loop metrics in their evaluation of different driving algorithms. In this section, we benchmark a large set of planners to analyze the alignment of closed-loop metrics with traditional distance-based open-loop metrics and the proposed PDMS.

\boldparagraph{Benchmark} Studying the relation of closed-loop and open-loop metrics necessitates access to a fully reactive simulator. To stay compatible with the dataset, we use the nuPlan simulator~\citep{Karnchanachari2024ICRA}, which enables simulation for privileged planners with access to ground-truth perception and HD map inputs. Similar to PDMS, nuPlan combines weighted averages and multiplied penalties in two official scores: the \textbf{open-loop score (OLS)} aggregates displacement and heading errors with a multiplied miss-rate, and the \textbf{closed-loop score (CLS)} implements similar metrics from~\secref{sec:3}. Including PDMS, all metrics are in $[0,1]$ with higher scores indicating better performance.

Due to the heavy computational requirements of closed-loop simulation, we evaluate on the \texttt{navmini} split. This is a new split we create for rapid testing, with 396 scenarios in total that are independent of both \texttt{navtrain} and \texttt{navtest} but filtered using the same strategy (\secref{sec:3_splits}) and hence similarly distributed. We note that nuPlan offers two kinds of background agents: reactive agents along lane centers based on the Intelligent Driver Model (IDM)~\citep{Treiber2000}, and non-reactive agents replayed from the dataset, which we employ unless otherwise stated. While reactive simulations of longer or dynamic lengths are generally desirable, \eg to evaluate long-term decisions, enabling this requires dedicated solutions to long-horizon simulation that are not currently available in nuPlan~\citep{Chitta2024ARXIV}. Therefore, we default to a fixed closed-loop simulation duration of $d=15\textrm{s}$, and a planning frequency of $f=10\textrm{Hz}$, which are the standard closed-loop simulation settings in nuPlan~\citep{Karnchanachari2024ICRA}.

\boldparagraph{Motion Planners} Open-loop metrics favor learned planners while rule-based approaches perform well in closed-loop evaluation in nuPlan~\citep{Dauner2023CORL}. We use a combination of both planner types in this experiment to cover different performance levels. In total, we include 37 rule-based planners with 2 constant velocity and 8 constant acceleration models, 15 IDM planners~\citep{Treiber2000}, and 12 PDM-Closed variants~\citep{Dauner2023CORL} which differ in hyperparameters for trajectory generation. For learned planning, we evaluate Urban Driver models~\citep{Scheel2021CORL} of 2 model sizes and 2 training lengths, and PlanCNN~\citep{Renz2022CORL} models with 15 input combinations of the BEV raster, ego status, centerline, and navigation goal. We train all models on $\{25\%, 50\%, 100\%\}$ of \texttt{navtrain} and an equally sized uniformly sampled subset of OpenScene, giving 114 learned planners. See the supplementary material for additional details. 

\boldparagraph{Results} The alignment between metrics is presented in \figref{fig:correlation} (\hyperref[fig:correlation]{a-e}). Compared to OLS, we consistently observe better closed-loop correlation for PDMS, in terms of Spearman's (rank) and Pearson's (linear) correlation coefficients. As shown in (\hyperref[fig:correlation]{a}), PDMS can capture the closed-loop properties of both learned and rule-based planners, whereas distance-based open-loop metrics show a clear misalignment. Decreasing the CLS duration in (\hyperref[fig:correlation]{b}) from $d=15$s to $d=4$s further raises the correlation of PDMS and OLS, as the simulation horizon more closely matches the open-loop counterparts. Interestingly, we observe a higher correlation of open-loop metrics in (\hyperref[fig:correlation]{c}) when reducing the planning frequency to $2$Hz. We expect a lower planning frequency to mitigate cumulative errors and enhance the controller's stability in simulation, leading to more precise trajectory execution. Moreover, we observe an increase in correlation for longer PDMS horizons in (\hyperref[fig:correlation]{d}), ranging from $h=2$s to $h=8$s. While predicting the future motion over 8s is challenging in uncertain scenarios, our results indicate the value of long horizons when evaluating motion planners. Lastly, replacing the non-reactive background agents with reactive IDM vehicles during closed-loop simulation in (\hyperref[fig:correlation]{e}) has little effect on the correlation, possibly due to the similar difficulty of both tasks~\citep{Dauner2023CORL}. 

\begin{figure}[t!]
    \centering
    \includegraphics[width=1.0\textwidth]{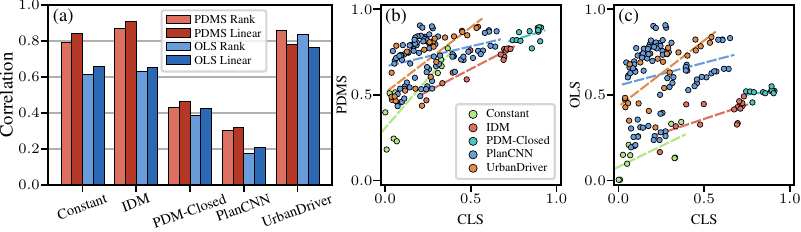}
    \caption{\textbf{Planner-Level Alignment of Metrics.} We report the correlation coefficients between open-loop metrics (OLS, PDMS) and the closed-loop score (CLS) for the five planner types considered in our study. The PDMS is better correlated to the CLS for every planner type.
    }
    \label{fig:correlation_2}
    \vspace{-0.0cm}
\end{figure}

The imbalanced distribution of different types of planners in our study may introduce biases into the overall correlations presented in \figref{fig:correlation}. To address this, we visualize the individual correlations of each planner type in \figref{fig:correlation_2}. The correlation values vary depending on metric range and variance of each planner type. Nevertheless, when examining each type individually, the PDMS is better correlated to the CLS than the OLS, and is always positively correlated.

\subsection{Analysis of the State of the Art in End-to-End Autonomous Driving}

In this section, we benchmark a collection of end-to-end architectures, which previously achieved state-of-the-art performance on existing open- or closed-loop benchmarks.

\begin{table*}[t!]
	\begin{center}
		\resizebox{\textwidth}{!}{
			\begin{tabular}{l|cccc|cc|ccc|c}
				\toprule
				\textbf{Method} & \textbf{Ego Stat.} & \textbf{Image} & \textbf{LiDAR} & \textbf{Video} & \textbf{NC} $\uparrow$ & \textbf{DAC} $\uparrow$ & \textbf{TTC} $\uparrow$ & \textbf{Comf.} $\uparrow$ & \textbf{EP} $\uparrow$ & \textbf{PDMS} $\uparrow$ \\
				\midrule
                Constant Velocity & \cmark & & & & 68.0 & 57.8 & 50.0 & 100 & 19.4 & 20.6\\
                Ego Status MLP & \cmark & & & & $93.0$ & $77.3$ & $83.6$ & $100$ & $62.8$ & $65.6$ \\
                \midrule
                LTF~\citep{Chitta2023PAMI} & \cmark & \cmark & & & $97.4$ & $\mathbf{92.8}$ & $92.4$ & $100$ & $79.0$ & $83.8$ \\
                TransFuser~\citep{Chitta2023PAMI} & \cmark & \cmark & \cmark & & $97.7$ & $\mathbf{92.8}$ & $92.8$ & $100$ & $79.2$ & $\mathbf{84.0}$ \\
                UniAD~\citep{Hu2023CVPR}  & \cmark & \cmark &  & \cmark & $97.8$ & $91.9$ & $92.9$ & $100$ & $78.8$ & $83.4$ \\
                PARA-Drive~\citep{Weng2024CVPR} & \cmark & \cmark &  & \cmark & $\mathbf{97.9}$ & $92.4$ & $\mathbf{93.0}$ & $99.8$ & $\mathbf{79.3}$ & $\mathbf{84.0}$ \\
                \midrule
                \textit{Human} & & & & & \textit{100} & \textit{100} & \textit{100} & \textit{99.9} & \textit{87.5} & \textit{94.8}  \\
				\bottomrule
			\end{tabular}}
	\end{center}
	\caption{\textbf{Navtest Benchmark.} We show the no at-fault collision (NC), drivable area compliance (DAC), time-to-collision (TTC), comfort (Comf.), and ego progress (EP) subscores, and the PDM Score (PDMS), as percentages. Relying on the ego status is insufficient for competitive results. While sensor agents improve, the gap to human performance highlights our benchmark's challenges.
    }
	\label{tab:sota}
    \vspace{-0.0cm}
\end{table*}  

\boldparagraph{Methods} As a lower bound, we consider the \textbf{(1) Constant Velocity} baseline detailed in \secref{sec:3_splits}. We include an \textbf{(2) Ego Status MLP} as a second "blind" agent, which leverages an MLP for trajectory prediction given only the ego velocity, acceleration and navigation goal. As an established architecture on CARLA, we evaluate our reimplementation of \textbf{(3) TransFuser}~\citep{Chitta2023PAMI}, which uses three cropped and downscaled forward-facing cameras, concatenated into a $1024\times256$ image, and a rasterized BEV LiDAR input for predicting waypoints. It performs 3D object detection and BEV semantic segmentation as auxiliary tasks. We then consider \textbf{(4) Latent TransFuser (LTF)}~\citep{Chitta2023PAMI}, which shares the same architecture as TransFuser but replaces the LiDAR input with a learned embedding, hence requiring only camera inputs. Moreover, we provide two state-of-the-art end-to-end architectures for open-loop trajectory prediction on nuScenes. \textbf{(5) UniAD}~\citep{Hu2023CVPR} incorporates a wide range of tasks, such as mapping, tracking, motion, and occupancy prediction in a semi-sequential architecture, which processes feature representations through several transformer decoders culminating in a trajectory planning module. \textbf{(6) PARA-Drive}~\citep{Weng2024CVPR} uses the same auxiliary tasks, but parallelizes the network architecture, such that the auxiliary task heads are trained in parallel with a shared encoder. Both UniAD and PARA-Drive use a BEVFormer backbone~\citep{Li2022ECCV}, which encodes the eight surround-view $1920\times1080$ camera images over four temporal frames into a BEV feature representation. Implementation details for all methods are provided in the supplementary material.

\boldparagraph{Results} We show our results on \texttt{navtest} in \tabref{tab:sota}. The Constant Velocity model is a lower bound, as the agent is used to identify trivial driving scenes excluded from the benchmark. The Ego Status MLP achieves a PDMS of $65.6$, showing the value of the acceleration and navigation goal for avoiding collisions and driving off-road. However, we observe a clear gap between agents relying solely on the ego status and those considering sensor data, in contrast to results on nuScenes~\citep{Li2024CVPR}. All sensor agents achieve a PDMS of over $83$, where TransFuser and PARA-Drive marginally perform best, with a PDMS of $84.0$. Surprisingly, the camera-only LTF achieves similar results ($83.8$). UniAD reaches a PDMS of $83.4$, which, together with PARA-Drive, do not surpass the performance of TransFuser and LTF, despite the need for more demanding training, \eg, 80 GPUs for 3 days to train PARA-Drive versus 1 GPU for 1 day for TransFuser on the \texttt{navtrain} split. Due to the definition of at-fault collisions, which discard certain rear-collisions into the ego vehicle, we suspect that surround-view cameras used by UniAD and PARA-Drive, and LiDAR input of TransFuser, are less important than the wide-angle front camera which is the only input of LTF. The $10$ PDMS discrepancy to the human operator demonstrates that \texttt{navtest} poses challenges even to well-studied end-to-end architectures. Specifically, the drivable area compliance (DAC) and ego progress (EP) subscores remain the most challenging. Notably, EP cannot be solved purely by human imitation, given that the maximum progress estimate used for normalization is based on a privileged rule-based motion planner. Interestingly, all agents achieve near-perfect comfort scores, indicating that smooth acceleration and jerk profiles are learned naturally from human imitation.

\begin{table*}[t]
    \centering
    \resizebox{\textwidth}{!}{
    \begin{tabular}{c|c|l|cc|ccc|c}
        \textbf{Config} & \textbf{Parameter} & \textbf{Setting} & \textbf{NC} $\uparrow$ & \textbf{DAC} $\uparrow$ & \textbf{TTC} $\uparrow$ & \textbf{Comf.} $\uparrow$ & \textbf{EP} $\uparrow$ & \textbf{PDMS} $\uparrow$ \\
        \toprule
        A1 & \multirow{3}{*}{Default config} & Seed 1 & $\mathbf{98.0}$ & $91.3$ & $\mathbf{94.2}$ & $100$ & $78.1$ & $83.3$ \\
        A2 & & Seed 2 & $97.7$ & $92.8$ & $92.8$ & $100$ & $79.2$ & $84.0$ \\
        A3 & & Seed 3 & $97.9$ & $\mathbf{93.0}$ & $93.1$ & $100$ & $\mathbf{79.3}$ & $\mathbf{84.4}$ \\
        \midrule
        B1 & \multirow{2}{*}{Ego status} & Goal only & $96.8$ & $91.9$ & $91.3$ & $98.6$ & $77.3$ & $81.8$ \\
        B2 & & Goal and velocity only & $96.7$ & $92.3$ & $91.0$ & $100$ & $77.8$ & $82.3$ \\
        \midrule
        C1 & \multirow{3}{*}{Camera FOV} & $60^\circ$ (1 camera) & $96.7$ & $90.2$ & $90.9$ & $100$ & $75.8$ & $80.3$ \\
        C2 & & $160^\circ$ (3 cameras) & $97.6$ & $91.4$ & $92.7$ & $100$ & $78.1$ & $82.8$ \\
        C3 & & $240^\circ$ (5 cameras) & $97.8$ & $92.5$ & $93.0$ & $100$ & $79.2$ & $84.1$ \\
        \midrule
        D1 & \multirow{3}{*}{LiDAR range} & F:$16$, B:$16$, L:$16$, R:$16$ & $96.9$ & $88.3$ & $91.2$ & $100$ & $74.6$ & $79.1$ \\
        D2 & & F:$64$, B:$32$, L:$32$, R:$32$ & $97.8$ & $92.7$ & $93.4$ & $100$ & $\mathbf{79.3}$ & $84.3$ \\
        D3 & & F:$64$, B:$64$, L:$64$, R:$64$ & $96.8$ & $90.3$ & $91.5$ & $100$ & $76.5$ & $81.0$ \\
        \midrule
        E1 & \multirow{2}{*}{Supervision} & No BEV segmentation  & $97.4$ & $90.5$ & $92.2$ & $100$ & $77.1$ & $81.6$ \\
        E2 & & No 3D detection & $97.8$ & $92.7$ & $92.9$ & $100$ & $79.2$ & $84.0$ \\
        \bottomrule
    \end{tabular}}
    \vspace{0.1cm}
    \caption{\textbf{TransFuser Ablations.} The default configuration, which obtains the best results, uses the navigation goal, velocity, and acceleration as ego status inputs. Its camera FOV is around 140$^\circ$ and LiDAR range is $32$m to the front (F), back (B), left (L), and right (R). It uses both auxiliary tasks.}
    \label{tab:transfuser}
    \vspace{-0.0cm}
\end{table*}

\boldparagraph{Analyzing TransFuser} In \tabref{tab:transfuser}, we compare several training settings for TransFuser. For the three training seeds in configs A1-A3, we observe a standard deviation of $\pm$ $0.56$ in PDMS, which is relatively small compared to variance among training seeds for closed-loop simulations in CARLA~\citep{Dosovitskiy2017CORL}. Further, unlike CARLA, NAVSIM is deterministic, and we obtain identical scores when repeating evaluations of a deterministic driving agent. Discarding velocity and acceleration (B1) lowers PDMS by $1.5-2.6$, whereas only removing the acceleration (B2) lowers the score by $1.0-2.1$. We conclude that while TransFuser benefits from the ego status, it is not purely relying on the kinematic state for planning. Next, only considering the front camera (C1) with a $60^\circ$ FOV leads to a small drop in almost all subscores, compared to our default setting of three cropped and concatenated images with a FOV of $140^\circ$. However, expanding the FOV with additional cameras does not result in substantially improved scores. Interestingly, restricting the LiDAR range to $16$m in all directions (D1), results in a score of $79$, which is lower than dropping LiDAR altogether (see LTF in \tabref{tab:sota}). Expanding the LiDAR range to $64$m in the forward direction (D2) or all directions (D3) does not provide significant improvements. We suspect that changes in the LiDAR range overly simplify or complicate the auxiliary 3D object detection and BEV semantic segmentation tasks, which operate in the LiDAR coordinate frame, hindering effective imitation learning. We check the impact of the auxiliary tasks by excluding them, where performance drops without BEV Segmentation (E1).

\begin{wrapfigure}{r}{0.36\textwidth}
    \vspace{-0.0cm}
    \centering
    \includegraphics[width=0.35\textwidth]{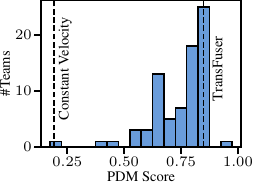}
    \caption{\textbf{NAVSIM Challenge.} 
    }
    \label{fig:challenge}
    \vspace{-0.3cm}
\end{wrapfigure}
\boldparagraph{CVPR 2024 NAVSIM Challenge} We organized the inaugural NAVSIM challenge which ran from March - May 2024. To ensure integrity, we used a private dataset and only gave participants access to sensor inputs, withholding all annotations. Competitors could submit their agent's trajectories to our leaderboard, where they were simulated and scored to obtain the PDMS. We received 463 submissions from 143 teams, of which 78 submissions were made publicly visible. We summarize their scores in \figref{fig:challenge}, relative to the constant velocity and TransFuser baselines from~\tabref{tab:sota}. The winning entry extended TransFuser and learned to predict proxy subscores for trajectory samples~\citep{Li2024ARXIV}, with a sampling strategy inspired by VADv2~\citep{Chen2024ARXIV}. These predicted subscores were weighted alongside a human imitation score to select the output plan. While the idea of sampling and scoring trajectories is well-known~\citep{Thrun2006JFR, Werling2010ICRA, Zeng2019CVPR, Casas2021CVPR, Dauner2023CORL}, it has recently been overlooked in favor of approaches which predict a single trajectory. This result prompts a reassessment of such methods. The team that placed second employed a vision language model (VLM) for driving, which is rapidly emerging as a sub-field in the AV literature~\citep{Sima2023ARXIV,Ma2023ARXIV,Yang2023ARXIV,Zhou2024ARXIV}. Several submissions attempted to reimplement or extend prior work on nuScenes such as UniAD~\citep{Hu2023CVPR} and VAD~\citep{Jiang2023ICCV}, but were unable to outperform the TransFuser baseline by the challenge submission deadline, given the significant engineering challenge and compute requirements. The diversity of the solutions on the leaderboard shows the potential of NAVSIM as a framework for pushing the frontiers of autonomous driving research. We aim to hold future competitions with more challenging data and metrics. Detailed competition results and statistics are provided in the supplementary material.

\begin{wraptable}{r}{5.cm}
    \vspace{-0.2cm}
    \centering
    \footnotesize
    \setlength{\tabcolsep}{4pt}
    \begin{tabular}{l|l}
    \toprule
    \textbf{Method} & \textbf{PDMS} $\uparrow$ \\
    \midrule
    TransFuser~\citep{Chitta2023PAMI} & $83.9 \pm 0.4$\\
    LTF~\citep{Chitta2023PAMI} & $83.5 \pm 0.6$\\
    Ego Status MLP & $66.4 \pm 0.9$\\
    \midrule
    Hydra-MDP~\citep{Li2024ARXIV} & $91.3$ \\
    Constant Velocity & $20.6$\\
    \bottomrule
\end{tabular}
    \caption{\textbf{Leaderboard 1.1.}}
    \label{tab:server}
    \vspace{-0.2cm}
\end{wraptable}

\boldparagraph{NAVSIM 1.1 Leaderboard} Due to the lasting interest after the challenge, we re-opened a public evaluation server using \texttt{navtest} as the evaluation split. The leaderboard encourages multi-seed submissions and includes reproducibility requirements for openly releasing code and model weights. We populated the leaderboard with 3 training seeds of our learned baselines, as shown in \tabref{tab:server}. For reference, we also include a single seed of the 2024 challenge winner~\citep{Li2024ARXIV} and constant velocity baseline. Further information is provided in the supplementary material and leaderboard webpage\footnote{\href{https://huggingface.co/spaces/AGC2024-P/e2e-driving-navsim}{{\color{ellisred}{\texttt{https://huggingface.co/spaces/AGC2024-P/e2e-driving-navsim}}}}}.

\section{Discussion}
\label{sec:5_discussion}

We present NAVSIM, a framework for non-reactive AV simulation. We address shortcomings of existing driving benchmarks and propose standardized but configurable simulation-based metrics for benchmarking driving policies. For accessibility, we provide challenging scenario splits and simple data curation methods. We show that our evaluation protocol is better aligned to closed-loop driving, benchmark an established set of end-to-end planning baselines from CARLA and nuScenes, and present the results of our inaugural competition. We hope that NAVSIM can serve as an accessible toolkit for AV researchers that bridges the gap between simulated and real-world driving. 

\boldparagraph{Need for Reactive Simulation} While we show improvements over displacement errors, several aspects of driving remain unaddressed by evaluation in NAVSIM. A high PDMS does not always imply a high CLS, since our framework does not consider reactiveness or the compounding accumulation of errors in closed-loop simulation. Moreover, as in CLS, rear-end collisions into the ego vehicle are currently not classified as "at-fault", resulting in little importance given to the scene behind the vehicle in NAVSIM. In the future, data-driven sensor or traffic simulation could alleviate these issues, once such methods mature and become computationally tractable. Given these limitations of the current framework, we strongly encourage the use of graphics-based closed-loop simulators, such as CARLA~\citep{Dosovitskiy2017CORL}, as complementary benchmarks to NAVSIM when developing planning algorithms.

\boldparagraph{Simplicity of Metrics} As a starting point, NAVSIM offers both interpretable open-loop subscores and a scalarizing function, which lets us provide a final score and ranking for participants in the challenge. In the future, multi-objective evaluation and other aggregation functions might be required. Moreover, closed-loop metrics also face problems, \ie, PDMS inherits several weaknesses of nuPlan's CLS. Both scores do not regard certain traffic rules (\eg, stop-sign or traffic light compliance) or concepts such as transit and fuel efficiency. In the future, we aim to improve the subscore definitions (\eg the at-fault collision logic) and add more subscores during aggregation. 

\boldparagraph{Call for Datasets} Certain limitations of the nuPlan dataset persist in NAVSIM, such as missing classes in the label space, minor errors in camera parameters, or noise in vehicle poses and 3D annotations. Our analysis might favor methods that are robust to such inconsistencies. In addition, the lack of road elevation data in our representation presents a challenge for integrating scenarios based on 3D map annotations. We aim to support more datasets in the future, and advocate for more open dataset releases by the community for accelerating progress in autonomous driving.

\section*{Acknowledgments}
This work was supported by the ERC Starting Grant LEGO-3D (850533), the DFG EXC number 2064/1 - project number 390727645, the German Federal Ministry of Education and Research: Tübingen AI Center, FKZ: 01IS18039A and the German Federal Ministry for Economic Affairs and Climate Action within the project NXT GEN AI METHODS. We thank the International Max Planck Research School for Intelligent Systems (IMPRS-IS) for supporting Daniel Dauner and Kashyap Chitta. We also thank HuggingFace for hosting our evaluation servers, the team members of OpenDriveLab for their organizational support, as well as Napat Karnchanachari and his team from Motional for open-sourcing their dataset and providing us the private test split used in the 2024 NAVSIM Challenge.

\medskip

\bibliographystyle{plainnat}
\bibliography{src/bibliography_long,src/bibliography,src/bibliography_custom}

\end{document}